\documentclass{vgtc}                          

\ifpdf
  \pdfoutput=1\relax                   
  \usepackage{graphicx}                
  \DeclareGraphicsExtensions{.pdf,.png,.jpg,.jpeg} 
\else
  \usepackage{graphicx}                
  \DeclareGraphicsExtensions{.eps}     
\fi%

\graphicspath{{figures/}{pictures/}{images/}{./}} 

\usepackage{microtype}                 
\PassOptionsToPackage{warn}{textcomp}  
\usepackage{textcomp}                  
\usepackage{mathptmx}                  
\usepackage{times}                     
\usepackage{cite}                      
\usepackage{tabu}                      
\usepackage{booktabs}                  

\usepackage{subcaption}
\usepackage{amssymb}

\onlineid{0}

\vgtccategory{Research}

\vgtcinsertpkg

\title{HypperSteer: Hypothetical Steering and Data Perturbation in Sequence Prediction with Deep Learning}

\author{Chuan Wang\thanks{e-mail: nauhcy@gmail.com}\\ %
        \scriptsize University of California, Davis %
\and Kwan-Liu Ma\thanks{e-mail:ma@cs.ucdavis.edu}\\ %
     \parbox{1.4in}{\scriptsize University of California, Davis}}

\abstract{
Deep Recurrent Neural Networks (RNN) continues to find success in predictive decision-making with temporal event sequences. Recent studies have shown the importance and practicality of visual analytics in interpreting deep learning models for real-world applications. 
However, very limited work enables interactions with deep learning models and guides practitioners to form hypotheticals towards the desired prediction outcomes, especially for sequence prediction. Specifically, no existing work has addressed the what-if analysis and value perturbation along different time-steps for sequence outcome prediction. 
We present a model-agnostic visual analytics tool, HypperSteer, that steers hypothetical testing and allows users to perturb data for sequence predictions interactively. We showcase how HypperSteer helps in steering patient data to achieve desired treatment outcomes and discuss how HypperSteer can serve as a comprehensive solution for other practical scenarios.

} 

\CCScatlist{ 
  {Human-centered computing - Visual Analytics}
}

\begin{document}


\firstsection{Introduction}

\maketitle

Recurrent neural networks (RNN) are widely used because of its remarkable performance in sequence modeling and prediction. RNNs are mostly used in the natural language processing (NLP) domain for text analysis but also demonstrates its power in other domains such as electronic health records (EHR) analysis \cite{Choi2016}, customer behavior analysis \cite{WangONM18}, and stock prediction \cite{Bao2017}.
In these domain analyses, people often need to inspect deep learning models and use model prediction to guide their decision making.  
Existing work has shown that visualization is useful in explaining the behavior of deep learning models like \cite{Strobelt2016LSTMVisAT}, \cite{ZhangWMLE2019Manifold} as summarized in the survey \cite{HohmanKPC2019}. 
However, very little work allows practitioners to interact with deep learning models and probe machine learning systems for model feedback.
The what-if tool (WIT) \cite{Wexler2019} from Google addresses similar problems in spirit. But WIT is not designed for sequence data analysis.
No existing work has focused on the what-if analysis and value perturbation for sequence prediction using deep learning models. For example, practitioners may have questions such as: How would changes to an instance at different time-steps affect a model's prediction? What patterns are learned by the model that can help improve the outcome of a sequence? How would a hypothesis affect the outcomes across all instances in all prediction classes?

This paper presents HypperSteer (\textbf{Steer}ing \textbf{Hyp}otheticals for data \textbf{Per}turbation), a model-agnostic visual analytics tool that allows users to perturb data instances (sequences) for real-time predictions interactively. To assist such perturbation, HypperSteer allows users to discover counterfactuals.
In the context of predictive models, given a test instance and the model's prediction, a counterfactual instance describes the necessary change in input features that alter the prediction to a predefined output \cite{molnar2019}. 
Examples of counterfactual instances resulting in a different prediction class can give powerful insights into what changes are necessary to change a sequence’s outcome.  
Furthermore, HypperSteer guides users towards reasonable hypothetical conditions using partial dependence (DP) analysis. HypperSteer also serves as a method to explain the predictions of RNN models. The contribution of this research is summarized as follows:
\vspace{-5px}
\begin{itemize}
    \item An interactive visualization system for what-if analyses in sequence prediction tasks. The approach allows users to conduct value-level feature perturbation and instantly predicts the outcome for sequential instances. 
    \vspace{-5px}
    \item A visual analytics approach that visualizes data evidence to guide users in data perturbation and hypothesis making towards desired prediction results. The data evidence is visualized at two levels: counterfactuals are visualized on the individual instance level, and features' partial dependence is summarized across entire prediction classes on the global level.
\end{itemize}

\section{Related Work}
HypperSteer relates to a large category of work that attempts to understand machine learning models using visualization.
LSTMVis \cite{Strobelt2016LSTMVisAT} visualizes the hidden state dynamics of RNNs by a parallel coordinates plot. To match similar top patterns, LSTMVis allows users to filter the input range and check hidden state vectors from heatmap matrices. For natural language processing tasks, Ming et al.~\cite{Ming2017} also employ a matrix design. Combining hidden state clusters and word clusters, Ming et al.~\cite{Ming2017} design a co-clustering layout, which links cluster matrices and word clouds.

As a subgroup, a number of black-box approaches focus on interpreting machine learning models by correlating data evidence to prediction outcomes. 
GAMut \cite{hohman2019} targeting the interpretation of generalized additive models is similar to HypperSteer in the idea of hypothesis testing. Likewise, iForest \cite{Zhao2019iForest} is designed for the what-if analysis based on random forest models.
RetainVis \cite{Kwon2018} facilitates users with the what-if prediction function using an approximated model to mimic the RNN model's performance around local instances. But RetainVis has a limited perturbation function that users can't perform value level predictions to test the actionable hypotheses.

The What-If Tool \cite{Wexler2019} is one of the closest creations aiming to test hypotheticals with model performance accessibility and help users understand model predictions. Similarly, ModelTracker \cite{amershi2015} enables flexible interaction to help model performance analysis and debugging. 
Prospector \cite{Krause2016Prospector} also provides interactive visualization to see model predictions responding to feature value perturbation.  
HypperSteer differs from these methods in many aspects. Unlike these methods, which are designed mostly for machine learning engineers or model developers, HypperSteer is designed for end-users like domain experts with end-user-friendly design and keeps visualization on the data level. Besides, HypperSteer is designed for sequence analysis, which is fundamentally different from these methods due to the increased time channel. 


\section {Design Consideration}
Due to the complexity of the what-if analysis for sequence predictions, the design requirements can be multitudinous. We summarize them into four design goals:

\textbf{DG1: Facilitate interactive perturbation on sequential data for hypotheses testing.}  HypperSteer should provide users with the ability to interact with a trained RNN model. Because the target users of HypperSteer are end-users who may not have a machine learning or computer science background, HypperSteer should use an easy-to-interpret graphic design for examining feature values on all time-steps. For users to test their hypotheses, the system design should facilitate intuitive and flexible interactions for users to change the feature values of any instance-of-interest at any time-steps, and the change can be applied to one or more feature dimensions.

\textbf{DG2: Facilitate accurate and real-time outcome inference in a model-agnostic manner.} 
Users should be able to interact with the model for outcome inference without knowing the model internals, such as the structure or layer outputs of the RNN model. This increase the interpretability of end-to-end model behavior. Therefore, users should be able to conduct model inference with any data instance and get instant feedback of the prediction result as they test their hypothesis. Different from many approaches that use approximate models to improve the computation speed in the inference, HypperSteer should target using the original RNN models for real-time inferences to reduce possible errors produced in the analysis procedures. 

\textbf{DG3: Guide users in data perturbation.} 
The requirements from users with hypotheses they want to test can be satisfied by the above design goals. However, because of the temporal dimension and the huge feature/value space, there are users who don't have clues to form hypotheses or don't know what degree of perturbation to make. Therefore, HypperSteer should guide users to 1) focus on important features and time-steps, and 2) form hypotheticals that can alter the prediction of a particular data instance to a predefined output. The fulfillment of these requirements also helps in the understanding of a model's local behavior around the instance-of-interest.

\textbf{DG4: Guide users to understand model predictions for entire classes.} 
Because deep learning models like RNNs are computed based on a populous distribution of instances, the understanding of models should not only base on their prediction behavior for individual instances. HypperSteer should let users understand model predictions by providing evidence from the entire prediction classes instead of cherry-picked examples.

\section{HypperSteer Functionalities}
This section introduces the functionalities of HypperSteer in correspondence with the design requirements in the last section. Overall, HypperSteer helps practitioners test hypotheticals with sequential data in prediction tasks and provides visual evidence that guides users to understand predictions and form hypotheticals. For users who have hypotheses to verify, the system design in subsection \ref{section:hypoTest} helps them interact with the pre-trained LSTM model and make predictions and provides counterfactual analysis results. The design introduced in subsection \ref{section:counterfact} and \ref{section:pdplot} illustrates how HypperSteer can help users make hypotheses by providing evidence that changing the outcome of selected instances. 
I use the MIMIC dataset to introduce the functionalities in the examples in the following subsections. A bidirectional LSTM model is trained and tested using the MIMIC dataset. The model uses patient medical records of the first 48 hours from admission to predict patients' mortalities (dead/alive) after 48 hours. This dataset contains more than 14K patients and 37 features at each time-step.

\subsection{Interactive Value-Level Hypotheses Testing}
\label{section:hypoTest}
Unlike machine learning researchers who pay more attention to the entire dataset's overall prediction performance, domain experts and practitioners usually focus on improving the outcome of individual instances. For example, in the MIMIC dataset analysis, clinical experts show more interest in improving the treatment outcomes for individual patients. Particularly, one of the most important goals is to find ways to ``flip'' the outcomes of certain patients-of-interest in the ``dead'' class. HypperSteer presents two linked modules to support the essential requirements in such analyses, as introduced in the following subsections.

\begin{figure*}
    \centering 
    \includegraphics[width=\textwidth]{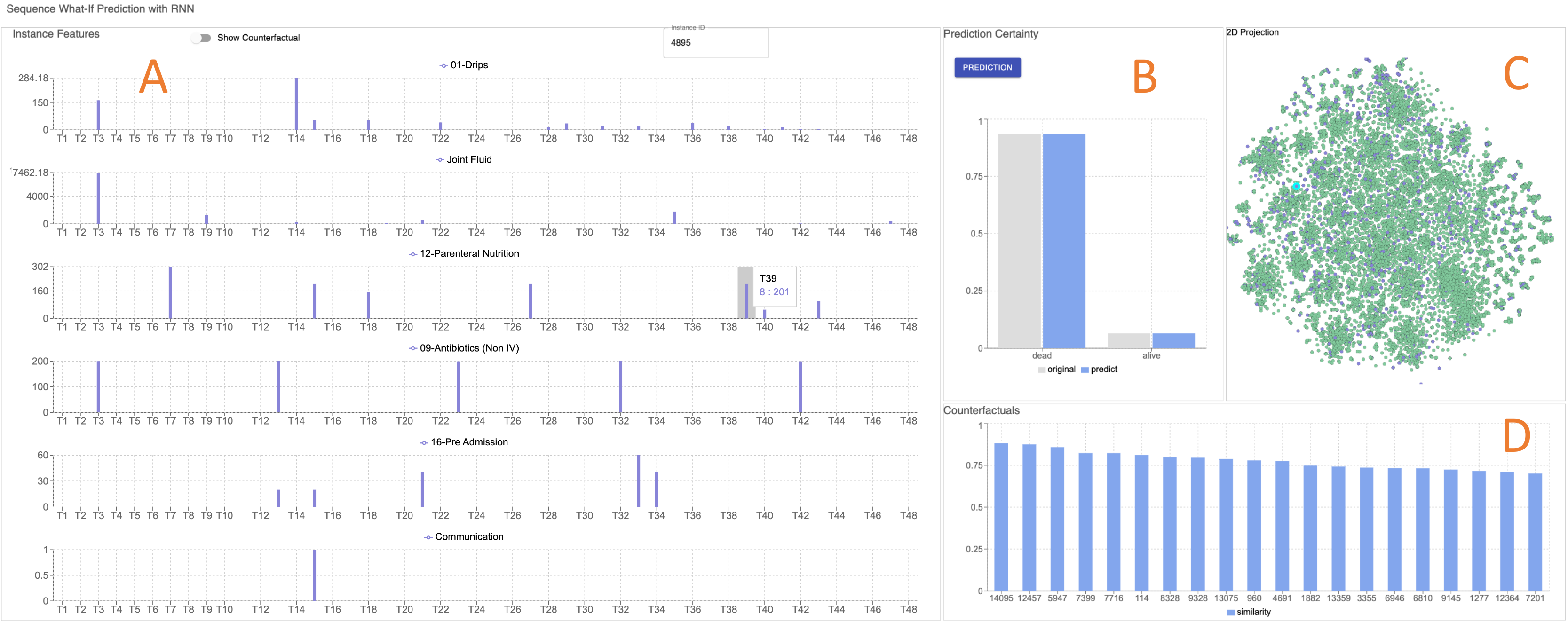}
    \caption{Hypotheses testing for individual instances.}
    \vspace{-5px}
    \label{fig:chap7-system}
\end{figure*}

\subsubsection{Instance Exploration}
As shown in Figure \ref{fig:chap7-system} (c), HypperSteer provides users with a 2D t-SNE view of the entire dataset.
Each dot represents one instance, which is a patient in the MIMIC scenario. 
The green color represents the negative class (dead in MIMIC), and the purple color represents the positive class (alive in MIMIC). A user can select a patient-of-interest by inputting the patient ID in the text field on the top or select a patient from the 2D projection view. When the user selects a patient, the corresponding point is highlighted in the 2D projection view, as shown in Figure \ref{fig:chap7-system} (c).

\subsubsection{Interactive Feature Value Perturbation and Prediction}
\begin{figure}[hbt]
    \centering
    \begin{subfigure}{\columnwidth}
        \centering
        \includegraphics[width=\columnwidth]{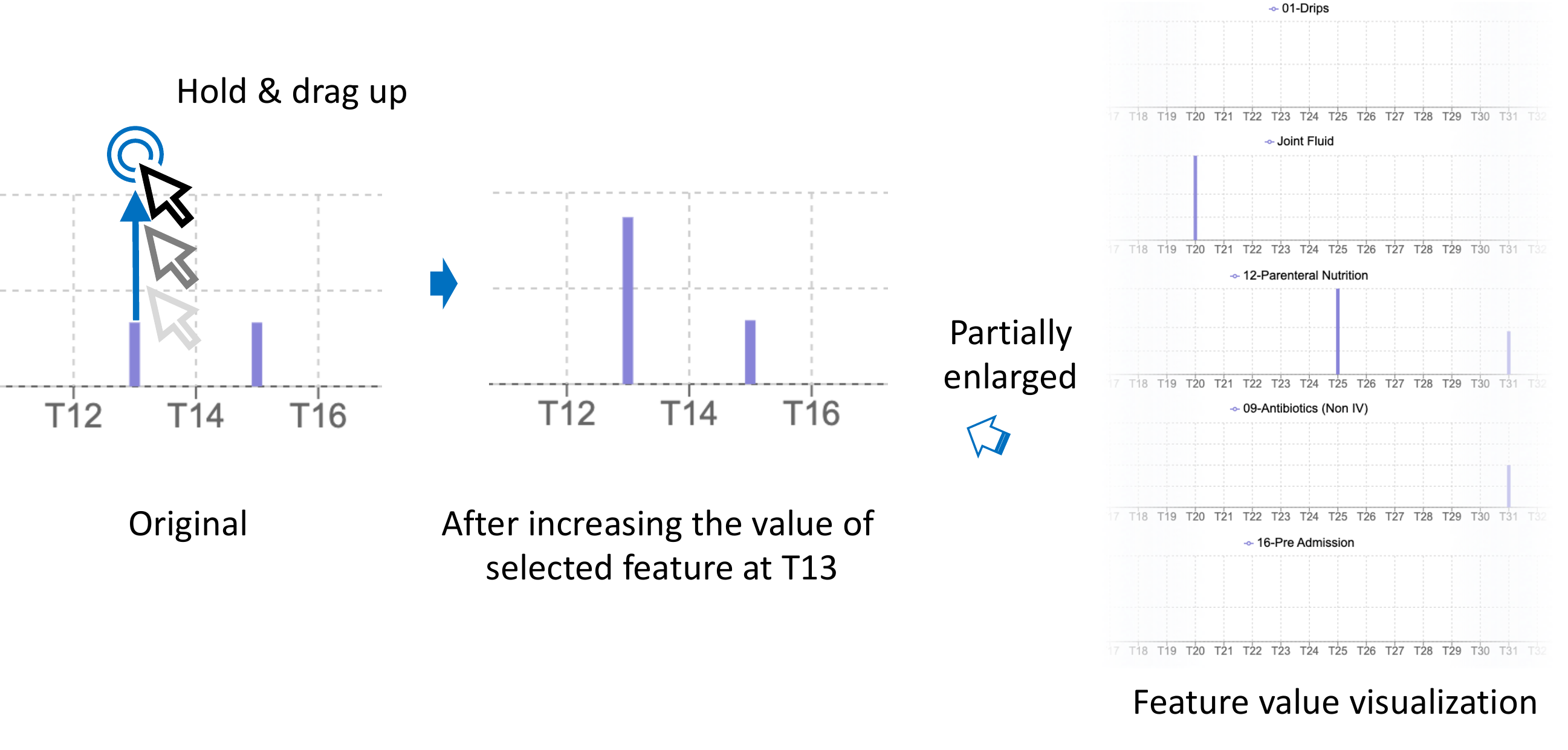}
        \caption{Interactive visualization of feature values.}
        \label{fig:feature-values-5}
    \end{subfigure}
    \begin{subfigure}{\columnwidth}
        \centering
        \includegraphics[width=\columnwidth]{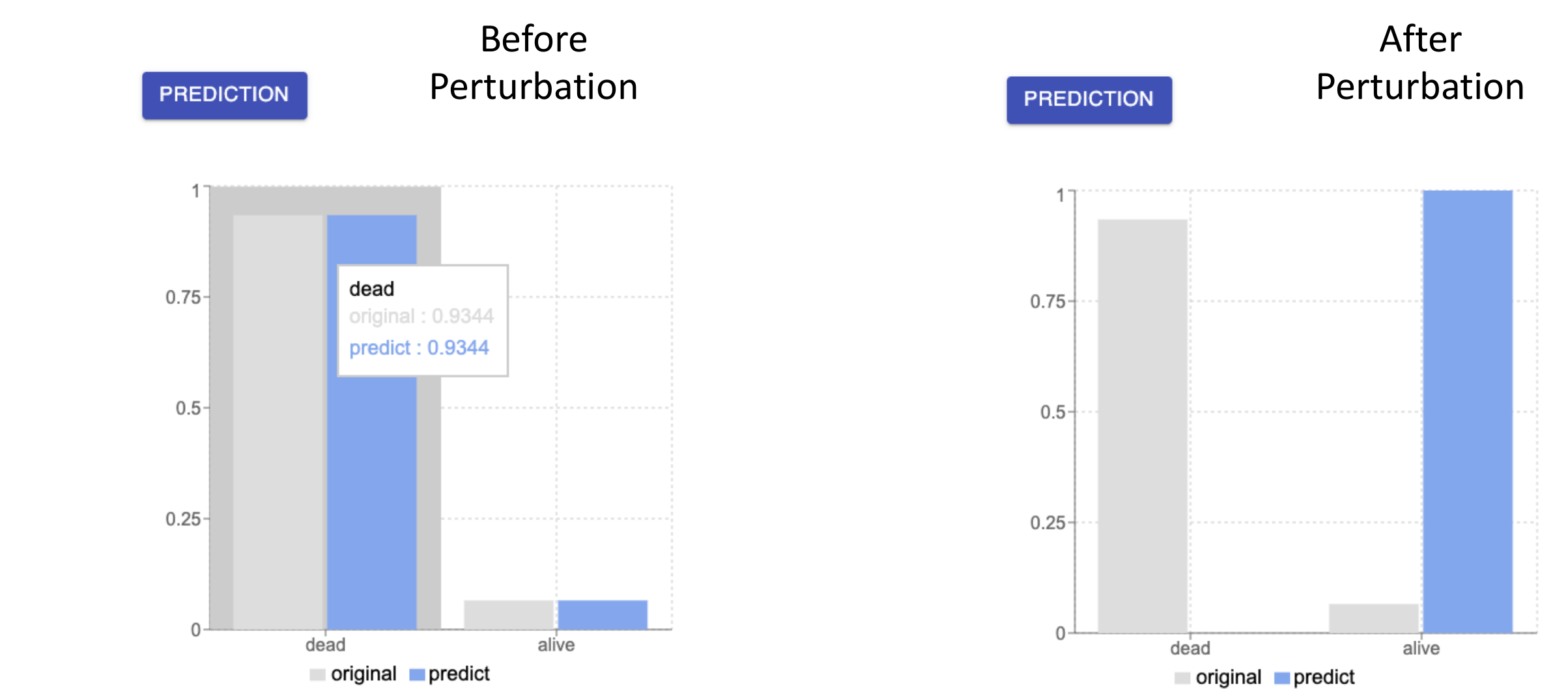}
        \caption{Prediction results.}
        \label{fig:feature-prediction}
    \end{subfigure}
    \caption{Visual components for hypotheses testing.}
    \vspace{-10px}
    \label{fig:hypo-test}
\end{figure}

To meet the requirements in DG1, we design an easy-to-interpret visualization for users to see the feature values of the selected instance. As shown in Figure \ref{fig:feature-values-5}, HypperSteer uses bar charts to show feature value at each time-step. Each single plot represents one feature with the name on the top. The horizontal axis represents time and the vertical axis represents the value of the feature. HypperSteer automatically narrows down the perturbation space (DG3) by visualizing the six features contributing most to the prediction in a contribution descending order from top to bottom \cite{Wang2020}. 
In the example shown in Figure \ref{fig:feature-values-5}, the patient doesn't have any ``Joint Fluid'' input through the period and all other features have sparse values that change over time. 

To adjust the visualization to match a hypothesis, a user can perturb feature values by modifying any selected feature' bar lengths at desired time-steps. When hovering the bar charts, HypperSteer provides real-time tool-tips to show bar values during the interaction. After confirming the visualization matches a hypothesis, the user can click the prediction button to make an inference with the pre-trained LSTM model (DG2), as shown in Figure \ref{fig:feature-prediction}. The instant feedback on predictions helps users confirm how changes to an instance at different time steps affect a model's prediction.

Figure \ref{fig:feature-prediction} shows the visualization of the prediction result before (left) and after (right) probing the feature values of a patient. HypperSteer uses two bars side-by-side to show the original and perturbed prediction results in a light gray and light blue color, respectively. Bar heights represent the probabilities of the corresponding class. In the example shown in Figure \ref{fig:feature-prediction}, the patient was originally dead, but after perturbing the medical inputs, the prediction results flipped to ``alive'', which means the LSTM model consider the perturbation is greatly correlated to a successful cure of the patient. However, how to find such a perturbation is a remaining question. HypperSteer uses visualizations to help users understand why models make predictions based on the data and guide users to form hypotheses for users who don't have clues of an applicable hypothesis. We introduce more details in the following sections.


\subsection{PD Plots: Global Evidence for Perturbation}
\label{section:pdplot}
It's essential for HypperSteer to guide users in data perturbation to achieve the desired outcomes. For instance, in the MIMIC data analysis, users would want to know what feature, at what time-step, and how to probe feature values to improve a patient's outcome. Understanding how models capture the dataset's underlying patterns can help users make such perturbation (DG3.)
HypperSteer visualizes the RNN model behavior across entire prediction classes to help users understand why a model makes predictions \textbf{based on a populous distribution of feature values} (DG4.)

Users can select a feature in the feature visualization view (Figure \ref{fig:chap7-system} A) by clicking the corresponding bar chart. HypperSteer calculates the prediction results at different values of the selected feature for all instances in the prediction class. 
HypperSteer automatically narrows down the temporal perturbation space (DG3) by  visualizing the partial dependence (PD) plot of the top contributing time-steps, using the feature attribution technique introduced in the case study section in \cite{Wang2020}. 

\begin{figure}[hbt]
    \centering
    \begin{subfigure}{0.49\columnwidth}
        \centering
        \includegraphics[width=\columnwidth]{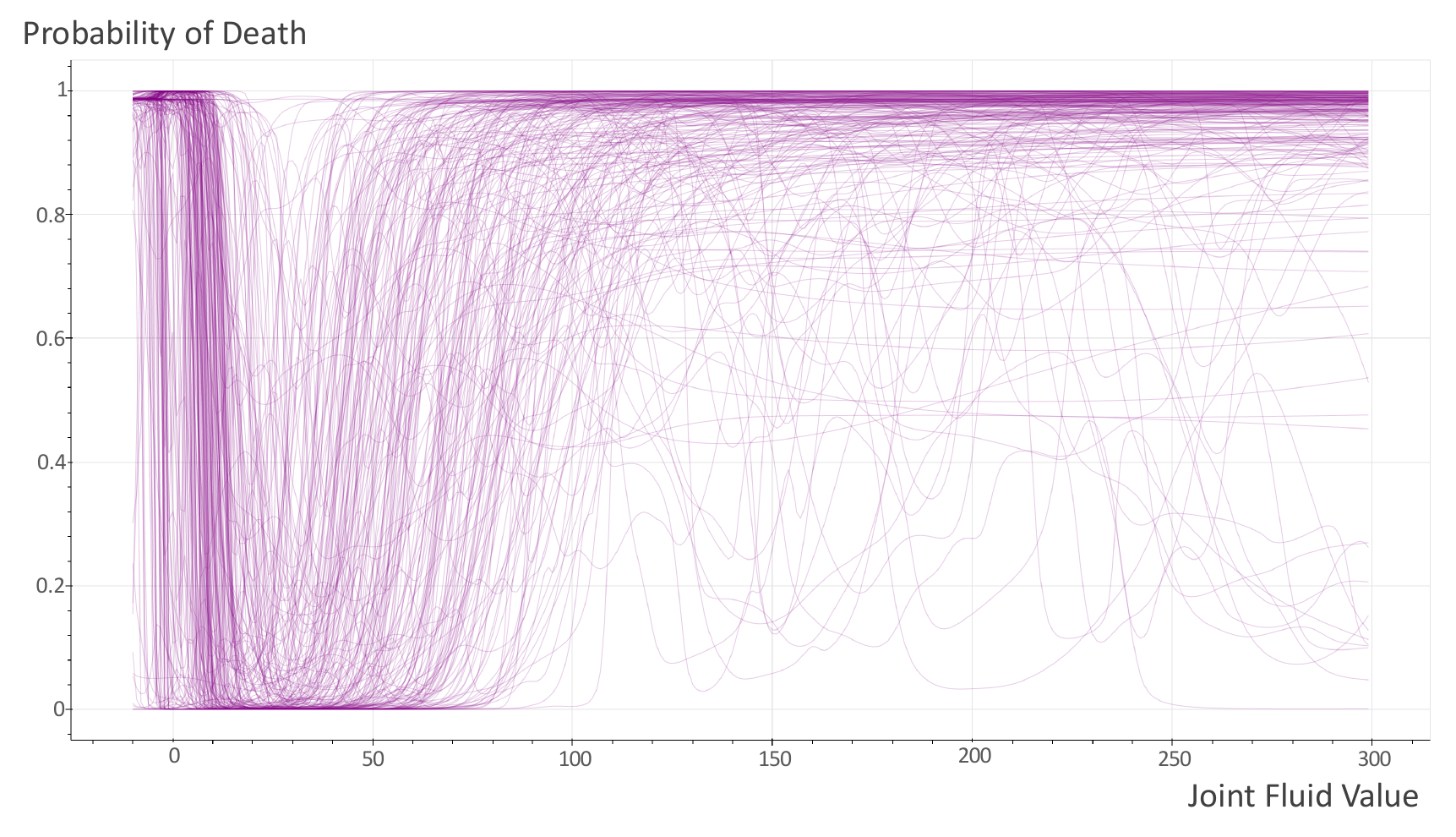}
        \caption{An example of partial dependence curves for a group of instances.}
        \label{fig:pdplot-raw3}
    \end{subfigure}
    \begin{subfigure}{0.49\columnwidth}
        \centering
        \includegraphics[width=\columnwidth]{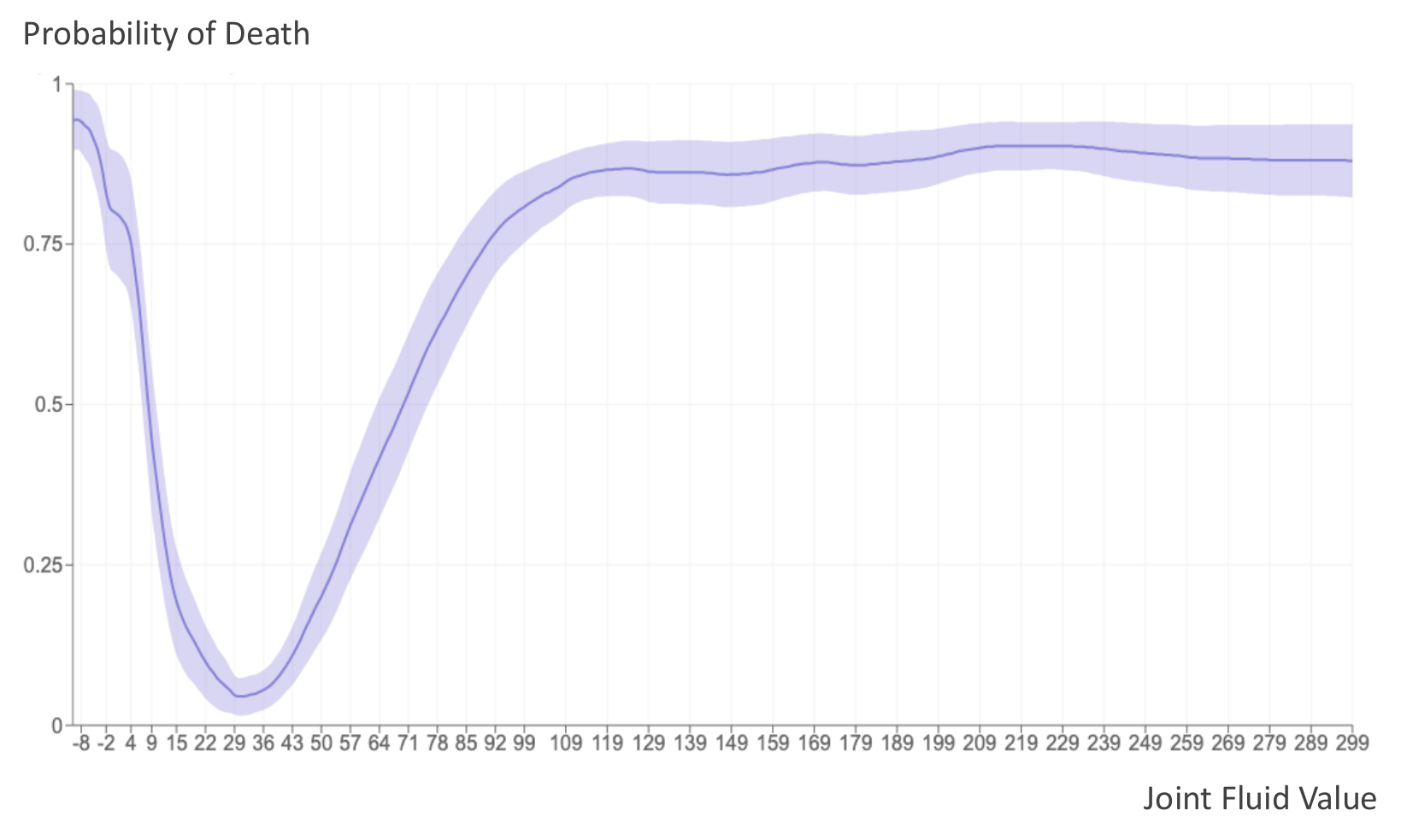}
        \caption{Visual summary of the overall trend in the left Figure with noise removal.}
        \label{fig:pdplot3}
    \end{subfigure}

    \caption{Partial dependence splines before and after summarization.}
    \vspace{-10px}
    \label{fig:pdplot-vis}
\end{figure}

\begin{figure}[ht]
    \centering 
    \includegraphics[width=0.8\columnwidth]{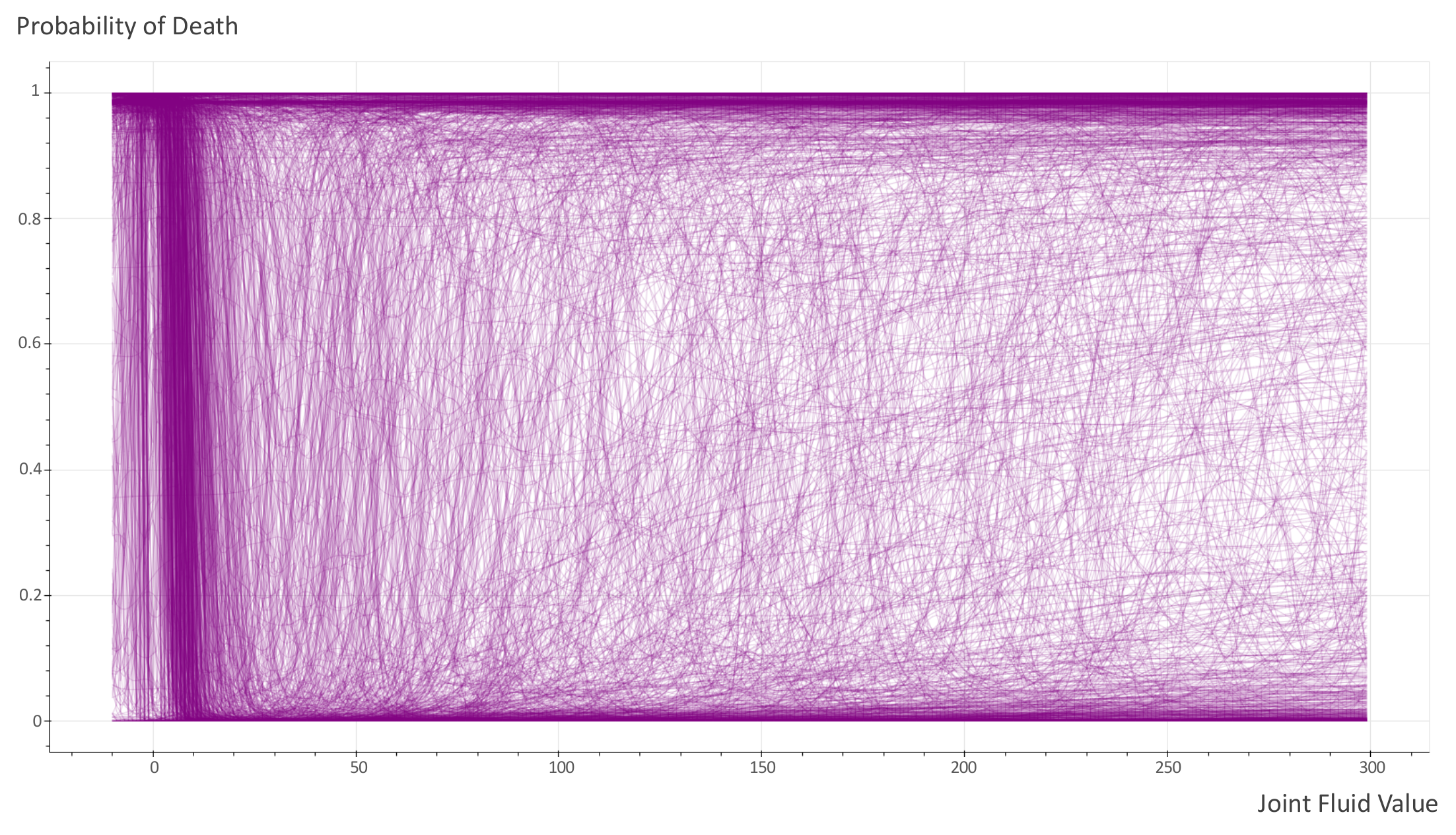}
    \caption{Raw PD plot across the entire dead patient class.}
    \vspace{-8px}
    \label{fig:pdplot-raw-all}
\end{figure}

The example in Figure \ref{fig:pdplot-raw3} shows the raw partial dependence splines for the last two hours in the 48-hour range after admission. The splines represent a group of ``dead'' patients in the EHR analysis. 
The horizontal axis represents different values for feature ``Joint Fluid'', and the vertical axis represents the probability that the RNN model predicts for ``dead''. Figure \ref{fig:pdplot-raw3} shows this group of patients share a similar trend in the prediction results when the feature values increase. 
However, when multiple patient groups have different trends, it's hard to tell their differences with the raw PD plot due to the overlapping noises. For example, Figure \ref{fig:pdplot-raw-all} shows the raw PD plot of the entire ``dead'' class. From such a plot, it's difficult to notice different patterns. Therefore, HypperSteer clusters the PD splines to distinguish different prediction patterns among instances. Figure \ref{fig:pdplot-raw3} is a showcase of one cluster in the ``dead'' patient class. Because the instance group size can be naturally uneven, clustering methods such as K-means and spectral clustering can generate misleading results in this scenario. Based on the performance, we choose the Agglomerative clustering method from multiple clustering methods for uneven cluster sizes such as DBSCAN and Gaussian mixtures. More results are discussed in Section \ref{section: chap7-result}.

Then HypperSteer summarizes the pattern in each cluster using one ribbon chart. As shown in Figure \ref{fig:pdplot3}, the horizontal and vertical axes have the same meaning as Figure \ref{fig:pdplot-raw3}. The ribbon chart illustrates the prediction result change over different feature values with a confidence interval shown as the ribbon. The line in the middle of the ribbon represents the mean value.
In statistics, the confidence interval (CI) gives an estimated range of values which is likely to include the mean value. Here, Each CI value at one time-step is calculated with the data samples at all different feature values. The CI has an associated confidence level 
$\gamma$. The CI calculated with $\gamma$ represents the frequency or the proportion of possible CIs that contain the mean value. For example, if $\gamma = 95\%$ then it's $95\%$ certain the visualized CI range contains the true mean of the values \cite{dekking2005}.

The visualization in Figure \ref{fig:pdplot-vis} shows the probability of death suddenly drops to almost zero when the feature values increase from minus ten to ten. When the values continue to grow to around 50, the probability of death increase again. This observation indicates that controlling the feature value to a range of zero to ten on the time-steps is likely to flip the mortality prediction result for this group of patients.

\subsection{Counterfactuals: Local Evidence for Perturbation}
\label{section:counterfact}
For individual instances, HypperSteer visualizes the counterfactuals to help users understand the differences that distinguish counterfactual pairs from two prediction classes locally. 
HypperSteer borrows Wexler et. al's methods \cite{Wexler2019} and computes the pairwise distances $\mathbb{D}$ between instances for the dataset at the back-end, and visualizes the counterfactuals ranked in a distance ascending order. 
As shown in Figure \ref{fig:chap7-system} D, the bar chart illustrates the instances of the top similarities in the other class respect to the selected instance. The horizontal axis represents different instance IDs, and the vertical axis represents the similarities between the counterfactuals and the selected instance-of-interest. 
For a uniform similarity measure across the dataset, the similarity $\delta$ of an instance $j$ and the instance of interest $i$ is calculated as follows:

\begin{equation}
    \delta_{i, j} = 1 - \tilde{D}_{i,j}
\end{equation}
\nolinebreak
$\tilde{D}$ represents the normalized distance between $i$ and $j$, where 
\begin{equation}
    \tilde{D}_{i,j} =  \frac{D_{i,j} - D_{min}}{D_{max} - D_{min}} \in [0, 1].
\end{equation}
\nolinebreak
$D$ is the distance between $i$ and $j$ \cite{Wexler2019}. Particularly, suppose the normalization is performed on the entire distance matrix $\mathbb{D}$. In that case, $\tilde{D}_{i,j}$ will be too small to visualize the counterfactuals because the most dissimilar pairs are too large for most datasets. HypperSteer calculates $D_{max}$ as the value of the $k^{th}$ percentile $P_{k}$ of the subset $\mathbb{D}_{n}$. $\mathbb{D}_{n}$ is the subset of $\mathbb{D}$ that is composed of all $n$ smallest distances of all instances. By default, HypperSteer uses $k = 80$ and $n = 1$ in the system. In other words, the maximum value that HypperSteer selects for the normalization guarantees $80\%$ ($k$) chances that a user selects one instance and there will be at least one ($n$) similar instance in the desired range.


\section{Result Showcase} 
\label{section: chap7-result}
Here we show the MIMIC data analysis results that HypperSteer guides users in data perturbation with both global and local evidence to flip patients' prediction outcomes from the dead to alive. 

\subsection{Global Evidence Guiding Data Perturbation}
\label{subsection: chap7-global}
In the MIMIC data analysis, the goal is to help clinical experts discover the perturbation that turns dead patient predictions alive.

\begin{figure}[hbt]
    \centering
    \begin{subfigure}{0.49\columnwidth}
        \centering
        \includegraphics[width=\columnwidth]{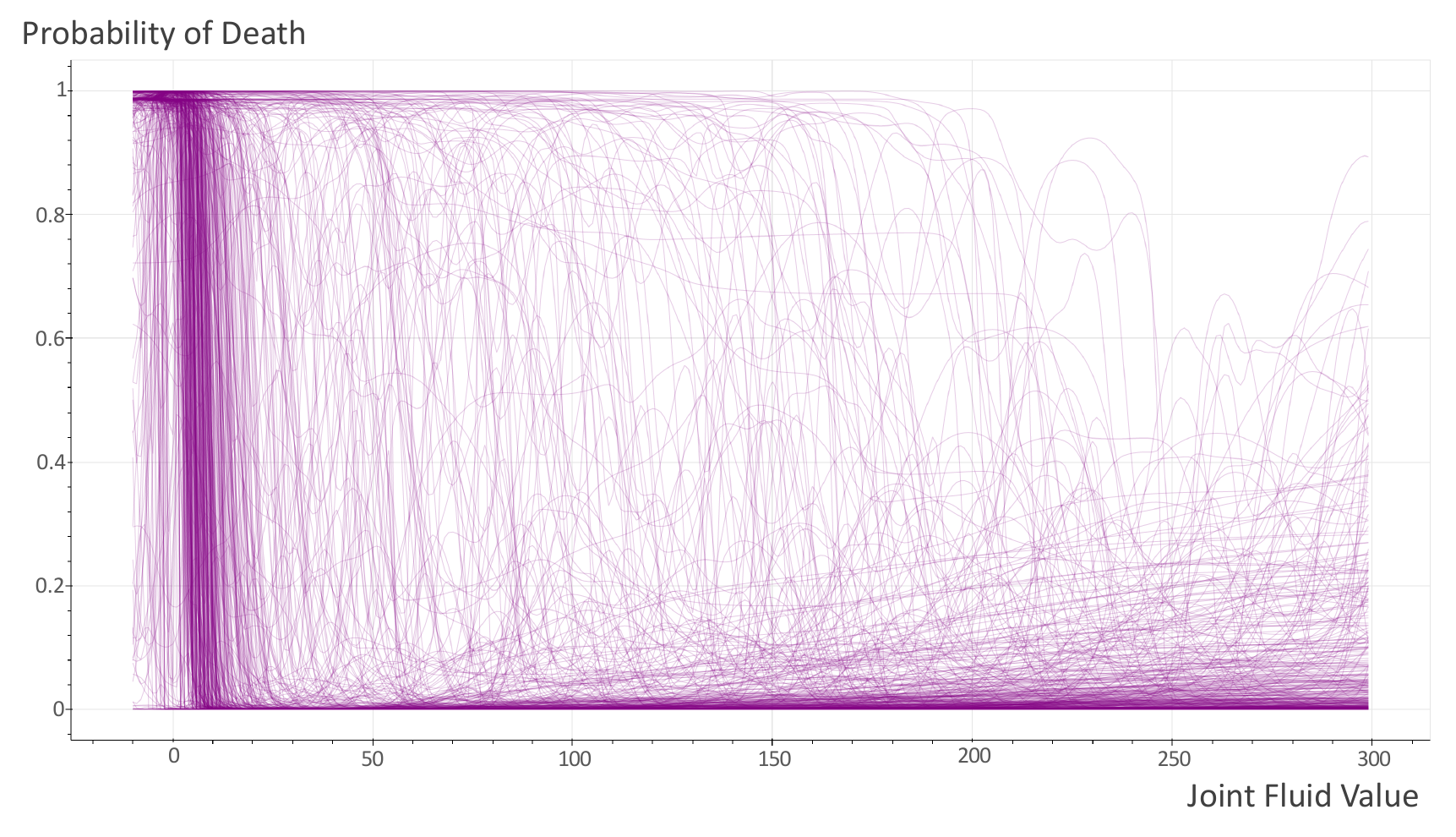}
        \caption{Cluster 1.}
        \label{fig:pdplot-cluster0}
    \end{subfigure}
    \centering
    \begin{subfigure}{0.49\columnwidth}
        \centering
        \includegraphics[width=\columnwidth]{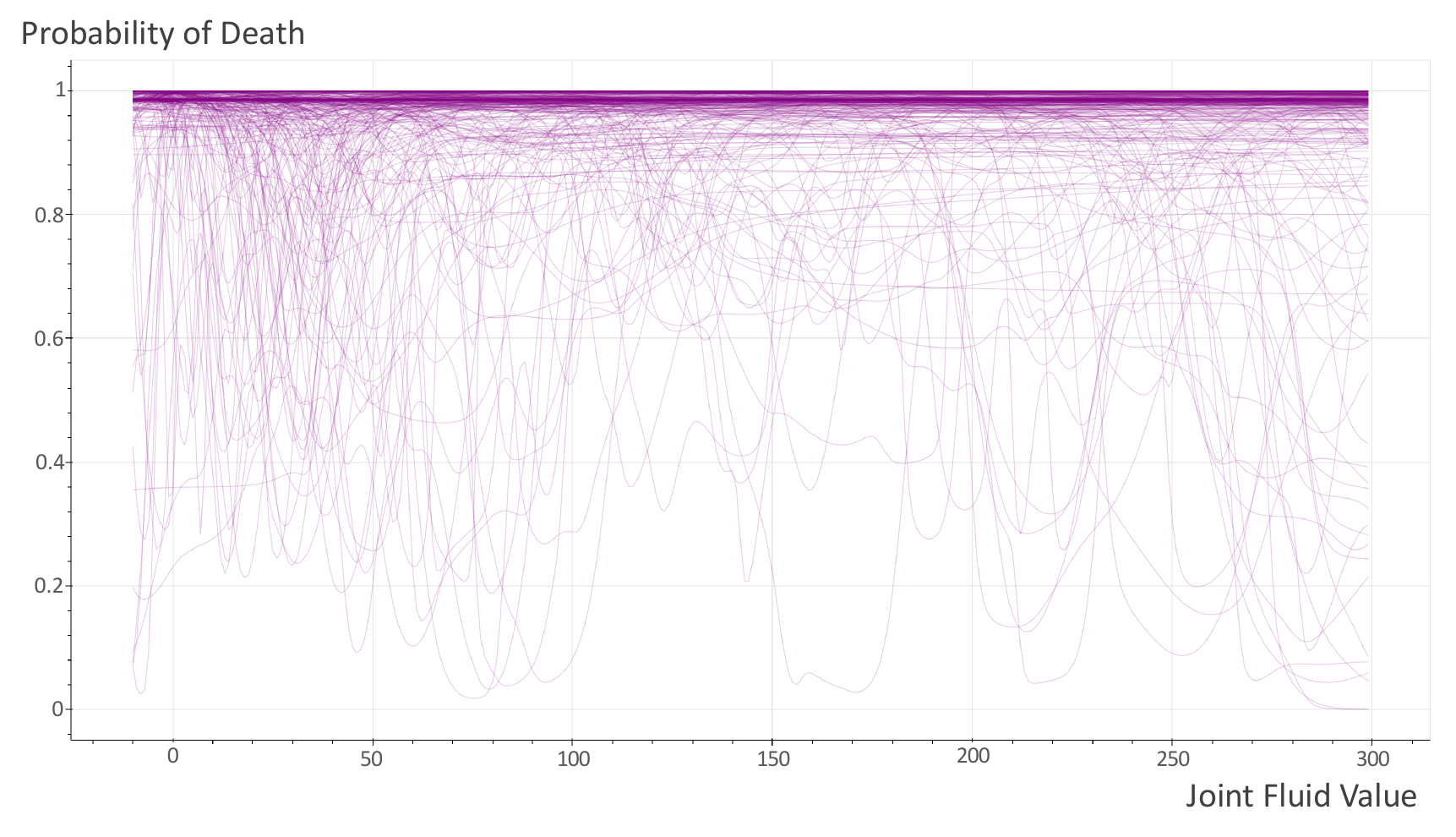}
        \caption{Cluster 2.}
        \label{fig:pdplot-cluster1}
    \end{subfigure}
    \centering
    \begin{subfigure}{0.49\columnwidth}
        \centering
        \includegraphics[width=\columnwidth]{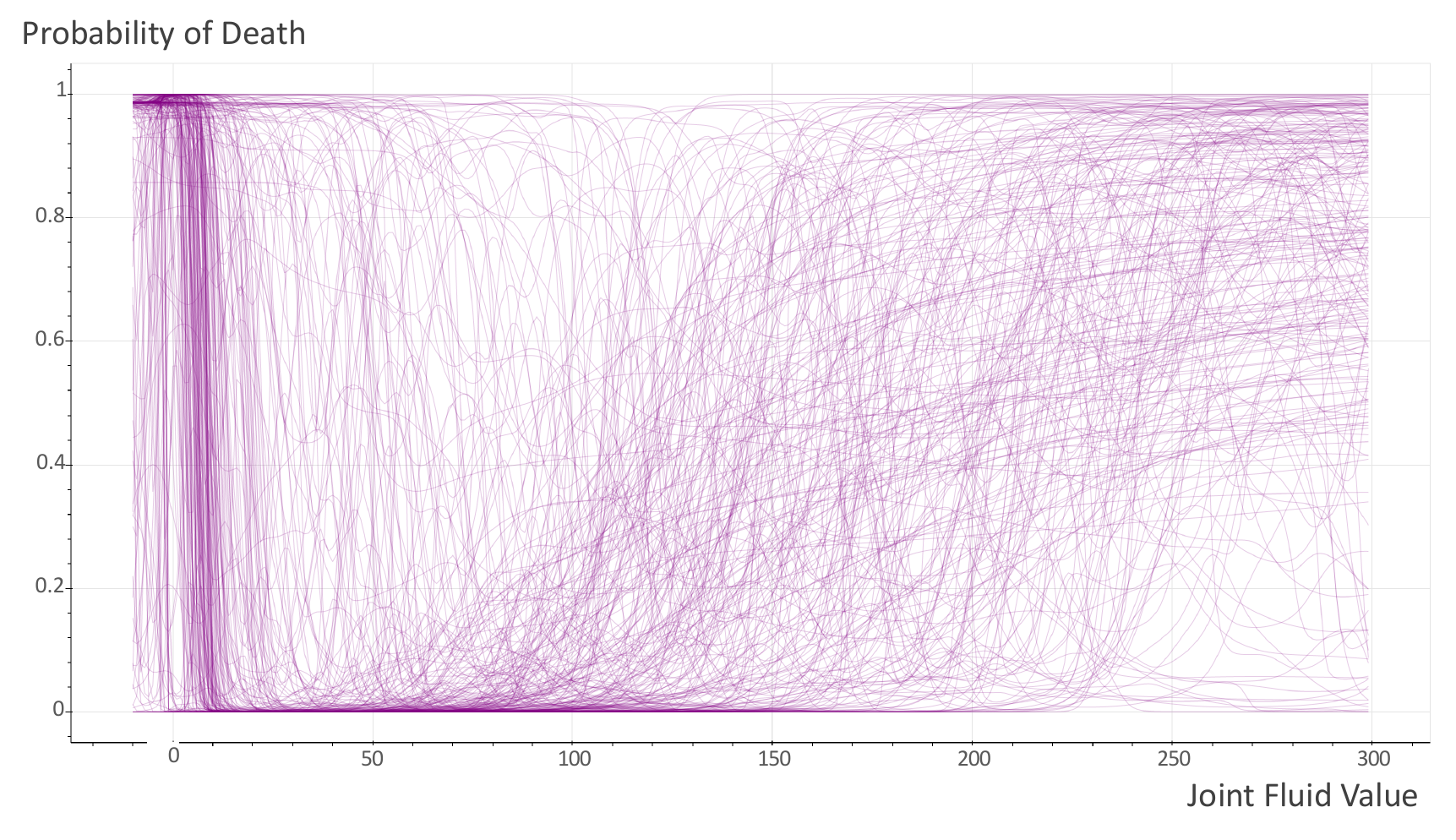}
        \caption{Cluster 3.}
        \label{fig:pdplot-cluster2}
    \end{subfigure}
    \begin{subfigure}{0.49\columnwidth}
        \centering
        \includegraphics[width=\columnwidth]{./fig/chap7/pdplot-raw3-label}
        \caption{Cluster 4.}
        \label{fig:pdplot-cluster3}
    \end{subfigure}
    \caption{Partial dependence spline clustering.}
    \vspace{-5px}
    \label{fig:raw-pd}
\end{figure}

\begin{figure}
    \centering 
    \includegraphics[width=\columnwidth]{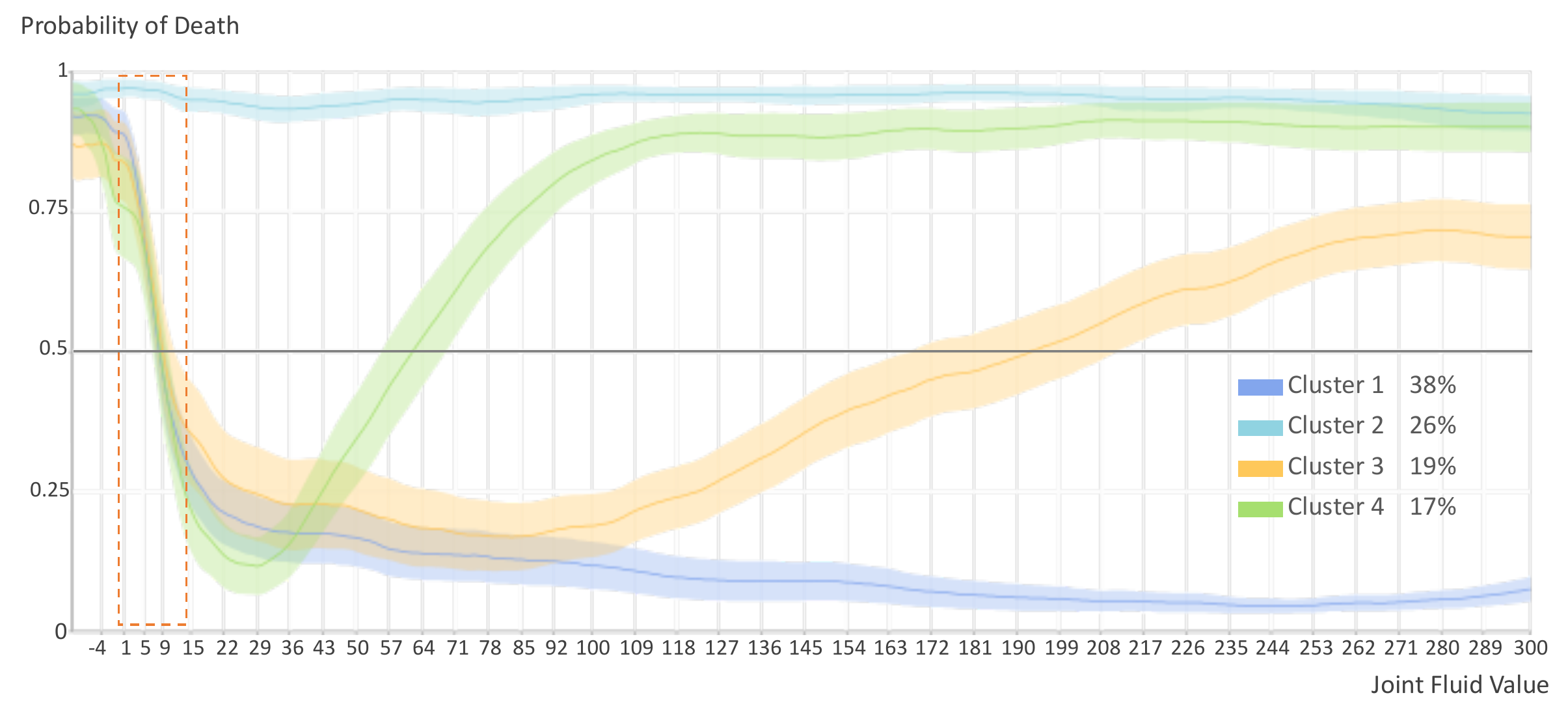}
    \caption{PD plot summarizing different values' influence on the model prediction across the entire prediction class.}
    \vspace{-10px}
    \label{fig:pdplot}
\end{figure}

Figure \ref{fig:raw-pd} shows the clustered partial dependence curves for the feature ``Joint Fluid'' of all patients (Figure \ref{fig:pdplot-raw-all}) in the ``dead'' class. The last two time-steps in the temporal sequences are computed as they have the top contributions to the prediction. The detected clusters have different prediction responses to different ``Joint Fluid'' values. HypperSteer calculates the clusters in the background and visualizes the summary with a PD plot (CI of 95\%), as shown in Figure \ref{fig:pdplot}.  In Figure \ref{fig:pdplot}, the horizontal and vertical axis represents feature value and predicted probability of death, respectively. The horizontal line is a separate of ``death'' and ``alive'' class, where the area above the line represents ``death'' and below ``alive.'' 

Each color represents one cluster corresponding with the clusters in Figure \ref{fig:raw-pd}. Cluster sizes are indicated as the percentage to the total size of the belonging class in the legend area on the bottom right. The visualization of the clusters exhibits a few characteristics: 1. cluster 2 is a patient group that their probability of death can hardly be changed over different ``Joint Fluid'' values.
2. The probabilities of death in clusters 3 and 4  decrease first and increase later, with different speeds. 3. cluster 1 shows an overall decrease in death probability.   
Interestingly, except for cluster 2, the other three clusters containing 74\% of the negative patients share a common pattern where the predicted probability drops dramatically around value 0-15 as shown in the orange rectangle. The phenomenon indicates that 74\% of dead patients if controlling their ``Joint Fluid'' values to 0-15, the probability of their death will decrease to the desired level and flip to alive. In other words, for a new patient of the same distribution as the original dataset, controlling their ``Joint Fluid'' values to this range has an around 70\% ($74\% * 95\%$) probability to turn a dead patient to alive under the guidance of the RNN model. 
The result helps users understand how model prediction performs over the entire dead patient class. The result also shows that controlling the top contributing feature ``Joint Fluid'' value in 0-15 has a great chance to flip the patient's death to alive under the RNN model's judgment, which provides the global evidence for data perturbations.

\subsection{Local Evidence Guiding Data Perturbation}
\label{subsection: chap7-local}
This section shows the analysis results for individual patients from the MIMIC dataset using a patient example. Similar analyses are conducted on more individual patients, and the results are coherent. 

\begin{figure}
    \centering
    \begin{subfigure}{\columnwidth}
        \centering
        \includegraphics[width=\columnwidth]{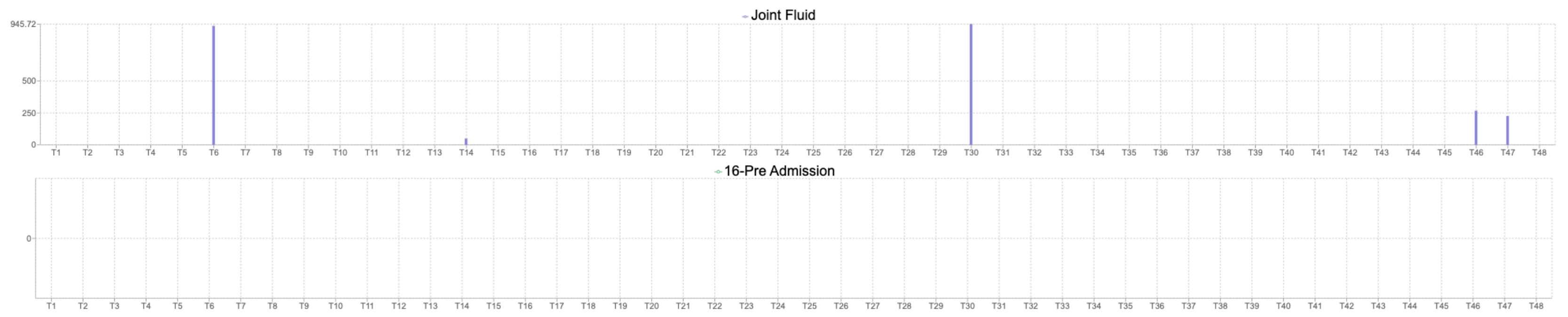}
        \caption{Patient 06712 (dead.)}
        \label{fig:p6712}
    \end{subfigure}
    \centering
    \begin{subfigure}{\columnwidth}
        \centering
        \includegraphics[width=\columnwidth]{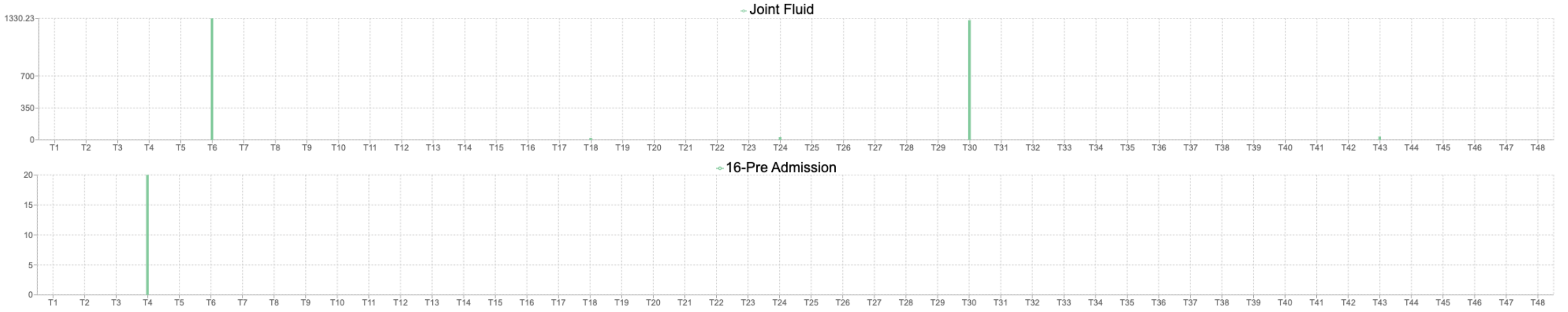}
        \caption{Patient 09009 (alive.)}
        \label{fig:p9009}
    \end{subfigure}
    \caption{A vanilla example of counterfactual patient pairs.}
    \vspace{-10px}
    \label{fig:patientpair1}
\end{figure}

A user first inputs a patient ID 06712 into the visualization view, and HypperSteer visualizes the value for features of top contribution from the feature visualization view. Patient 06712 (Figure \ref{fig:p6712}) is from the dead class. The user tries to understand what perturbation can possibly change the patient medication outcome to alive. The user selects patients by clicking the bar in the counterfactual view to review patients from the alive class, which is mostly similar to patient 06712. Figure \ref{fig:p9009} shows the counterfactual prototype, patient 09009, belonging to the alive class.  Comparing these two patients, the user notices a few characteristics.
Both 06712 and 09009 have highly sparse features, and only features with non-zero values are shown. And they both have high ``Joint Fluid'' values at T6 and T30. Different from 06712, 09009 has a ``16-Pre Admission'' value at T4. And 06712 has two ``Joint Fluid'' values around 250 at T46 and T47. Noticing the difference, the user decreases the ``Joint Fluid'' values at T46 and T47, increases the ``16-Pre Admission'' value at T4, and observes the change in the prediction of 06712. As a result, the decreasing the ``Joint Fluid'' value at T47 to an amount to 0-10 dramatically flips the prediction of 06712. Figure \ref{fig:hypo-test} shows the prediction result where the probability of alive increase from 0.1 (left) to 0.98 (right). This study shows that the counterfactual analysis can help users narrow down the perturbation space in both feature and time dimensions. The counterfactual analysis can also help identify the amount of perturbation needed to achieve the desired prediction outcome.

The visual analytics metrics in Subsection \ref{subsection: chap7-global} and \ref{subsection: chap7-local} both function as a guide for the user to interpret RNN models' prediction on the data level. Although the local counterfactual analysis may not guarantee to find a small number of necessary changes in the input features to alter the prediction for a particular instance, the case studies show a coherent result in many aspects. For instance, in the example of perturbing feature values of patient 06712, decreasing the ``Joint Fluid'' level at T46 won't affect the prediction result as much as T47, and so as changing the value of ``16-Pre Admission'' at T4. These results are coherent with the global pattern in Subsection \ref{subsection: chap7-global}, where perturbing the feature values at the last two time-steps causes the greatest reversal in the prediction.

\section{Discussion and Conclusion}
The counterfactual searching metric in this study discovers counterfactual prototypes. The merit of discovering counterfactual prototypes is that the discovered prototypes are always valid and real. However, this approach may discover zero counterfactuals if the distance threshold for calculating the counterfactual similarities is set to a small value for some datasets. Another way is to compute synthetic counterfactuals by optimizing the distance between the generated counterfactuals with regularization on the distribution and model prediction results, etc. In such scenarios, extra evaluation of computed counterfactuals may be required. 

In summary, we present HypperSteer, a visual analytics tool that helps users perturb sequence data towards the desired prediction outcome with RNN models. This tool enhances prescriptive visual analyses by allowing users to interactively modify the feature values at any time-step and provides real-time sequence inferences with the pre-trained RNN model. During the interactive data perturbation, the visualization in HypperSteer provides data evidence that guides users in making hypotheses, which helps users prescribe solutions in sequence prediction problems.


\bibliographystyle{abbrv-doi}

\bibliography{hyppersteer}
\end{document}